\documentclass{ieeeaccess}

\usepackage{amsmath,amssymb,amsfonts}
\usepackage{algorithmic}
\usepackage{graphicx}
\usepackage{textcomp}
\usepackage{booktabs}
\usepackage{multirow}
\usepackage{url}

\graphicspath{{figures/}}

\pdfmapfile{+t1-formata.map}
\pdfmapfile{+t1-giovannistd.map}
\pdfmapfile{+t1-helvetica.map}
\pdfmapfile{+t1-times.map}

\usepackage[backend=biber,style=ieee,citestyle=numeric,sorting=none,doi=false,isbn=false,url=false]{biblatex}
\usepackage[colorlinks=true,allcolors=blue]{hyperref}

\DeclareCiteCommand{\cite}
  {}
  {\printtext[bibhyperref]{\mkbibbrackets{\usebibmacro{citeindex}\printfield{labelnumber}}}}
  {\addcomma\space}
  {\usebibmacro{postnote}}

\usepackage{bm}
\makeatletter
\AtBeginDocument{\DeclareMathVersion{bold}
\SetSymbolFont{operators}{bold}{T1}{times}{b}{n}
\SetSymbolFont{NewLetters}{bold}{T1}{times}{b}{it}
\SetMathAlphabet{\mathrm}{bold}{T1}{times}{b}{n}
\SetMathAlphabet{\mathit}{bold}{T1}{times}{b}{it}
\SetMathAlphabet{\mathbf}{bold}{T1}{times}{b}{n}
\SetMathAlphabet{\mathtt}{bold}{OT1}{pcr}{b}{n}
\SetSymbolFont{symbols}{bold}{OMS}{cmsy}{b}{n}
\renewcommand\boldmath{\@nomath\boldmath\mathversion{bold}}}
\makeatother

\def\BibTeX{{\rm B\kern-.05em{\sc i\kern-.025em b}\kern-.08em
    T\kern-.1667em\lower.7ex\hbox{E}\kern-.125emX}}

\begin{document}
\history{Date of publication xxxx 00, 0000, date of current version xxxx 00, 0000.}
\doi{10.1109/ACCESS.XXXX.DOI}

\title{NAVIR: Neuromorphic Audio-Visual Speech Recognition for Robust Human--Robot Interaction on Edge Hardware}

\author{
\uppercase{Leonidas Delimpasis}\authorrefmark{1},
\uppercase{Panagiota Moraiti}\authorrefmark{2},
\uppercase{Antonis Porichis}\authorrefmark{3},
\uppercase{Panos Chatzakos}\authorrefmark{2,3}, and
\uppercase{Michail Karamousadakis}\authorrefmark{1}}

\address[1]{Plaixus Ltd., Spyrou Patsi 62, 11855, Athens, Greece}
\address[2]{Tech Hive Labs, 280 Kifisias Ave., 152 32 Halandri, Greece}
\address[3]{AI Innovation Centre, University of Essex, Little Abington, CB21 6GP Cambridge, U.K.}

\tfootnote{This work was carried out within the NAVIR project, which received funding from the European Union's Horizon Europe Research and Innovation Programme (dAIEDGE) under Grant Agreement No.\ 101120726.}

\markboth%
{L. Delimpasis et al: NAVIR: Neuromorphic AVSR for Robust Human--Robot Interaction on Edge Hardware}%
{L. Delimpasis et al: NAVIR: Neuromorphic AVSR for Robust Human--Robot Interaction on Edge Hardware}

\corresp{Corresponding author: Leonidas Delimpasis (e-mail: leonidasd@disroot.org).}

\begin{abstract}
Voice-controlled interaction in industrial settings is hampered by acoustic noise, which severely degrades audio-only speech recognition. Audio-visual speech recognition (AVSR) addresses this by fusing lip-motion cues with the audio stream, but state-of-the-art pipelines rely on three-dimensional convolutions, recurrent units, and attention modules that exceed the budget of typical edge devices. We present NAVIR, an end-to-end AVSR system targeting the BrainChip Akida neuromorphic processor, which natively supports only sequential two-dimensional convolutional inference. The pipeline factorises spatial and temporal encoding into separate AkidaNet-based modules: a per-frame visual encoder, a temporal video encoder, and a spectrogram audio encoder, fused by a lightweight predictor head and decoded by constrained beam search. Models are trained with connectionist temporal classification on noise-augmented audio and then fine-tuned with quantization-aware training. On the GRID benchmark, the quantized audio-visual model reaches 14.0\% word error rate (WER) under noise on the unseen-speaker split and 3.3\% WER on the overlapped-speaker split, against 22.5\% and 11.8\% for audio-only baselines, and it attains 98.6\% command accuracy at 1.5\% WER on a task-specific industrial-command corpus. Operation-count analysis indicates a 13-fold energy advantage of the spiking formulation over its artificial neural network counterpart at 27.6\% mean firing rate. On-board measurements show roughly 5-fold lower energy per inference than a Raspberry Pi central processing unit on the lip-reading model, and over 100-fold lower than a laptop graphics processing unit, while sustaining 14.5 inferences per second. To the best of our knowledge, this is the first complete multimodal AVSR pipeline running on neuromorphic hardware of this class.
\end{abstract}


\titlepgskip=-21pt

\maketitle

\section{Introduction}
\label{sec:introduction}

\PARstart{S}{peech} is one of the most natural modalities for human-machine interaction, yet its deployment in industrial environments remains limited because acoustic noise from machinery, ventilation and ambient activity sharply degrades the performance of audio-only automatic speech recognition (ASR). Workers therefore fall back on manual interfaces such as keypads, touchscreens or physical switches, which slow workflows and can compromise safety in tasks that require both hands free. The human perceptual system mitigates this fragility by integrating the visual appearance of the speaker's mouth, an effect captured most vividly by the McGurk illusion~\cite{Mcgurk_1976}. Audio-visual speech recognition (AVSR) systems aim to exploit the same redundancy and have been shown to deliver substantial robustness gains at low signal-to-noise ratios.

Modern high-accuracy AVSR pipelines, however, rely on three-dimensional convolutions, attention-based encoders, conformer back-ends or large transformer language models. These architectural choices implicitly assume GPU-class hardware and are difficult to deploy on the embedded, battery-powered or thermally constrained platforms found in mobile robotics. This creates a tension between recognition performance and deployment feasibility that is central to embedded multimodal speech interfaces.

Neuromorphic processors offer a promising route through this tension. Inspired by biological neural systems, they process sparse, event-driven binary spikes rather than dense floating-point activations, so that energy consumption scales with network activity rather than nominal compute capacity. The BrainChip AKD1000 is a commercially available neuromorphic system-on-chip that natively executes convolution-based spiking neural networks (SNNs) compiled via the Akida MetaTF/CNN2SNN toolchain. It connects to a host platform, in our case a Raspberry Pi 5, via PCIe, with total system draw in the 350--430\,mWh band over a five-minute inference session. Despite growing interest in neuromorphic computing for sensor-level event detection and gesture recognition, its application to multi-stream temporally structured tasks such as AVSR has not previously been demonstrated at the system level.

\textbf{Contributions.} The NAVIR project closes this gap with the following contributions.
\begin{itemize}
\item An end-to-end AVSR pipeline that satisfies the strict architectural constraints of the AKD1000 (no 3D convolutions, no recurrent layers, no attention, fixed quantization) by factorising spatial and temporal encoding into separate AkidaNet-based stages.
\item A constrained beam-search decoder that exploits the known grammar of the target vocabulary to guarantee grammatically valid output and reduce the search space at inference time.
\item Noise-robust training through aggressive UrbanSound8K augmentation, demonstrating consistent multimodal robustness gains over audio-only baselines on the GRID benchmark and on an internal industrial-command corpus.
\item A theoretical and practical energy analysis comparing our SNN against state-of-the-art (SOTA) artificial neural network (ANN) lip readers using the operation-count framework with Horowitz's 45\,nm CMOS energy constants~\cite{Horowitz_2014}, and against CPU- and GPU-based inference via on-board power measurements.
\item An interactive demonstration coupling the pipeline to a uFactory xArm 6 robotic arm, providing closed-loop validation from voice command to physical action on commodity embedded hardware.
\end{itemize}

The remainder of the paper is organised as follows. Section~\ref{sec:related} reviews neuromorphic SNN architectures, prior AKD1000 applications and the GRID lip-reading state of the art. Section~\ref{sec:method} describes the NAVIR architecture and decoding strategy. Section~\ref{sec:datasets} introduces the two evaluation corpora, and Section~\ref{sec:setup} the experimental setup. Recognition results are reported in Section~\ref{sec:results}, followed by the energy and power analysis in Section~\ref{sec:energy}. Section~\ref{sec:demo} presents the robot demonstration system, Section~\ref{sec:discussion} discusses limitations and future work, and Section~\ref{sec:conclusion} concludes. Appendix~\ref{app:av_slower} explains a hardware-mapping behaviour of the AKD1000 toolchain that affects the audio-video deployment, and Appendix~\ref{app:layers} details the layer composition of every module.

\section{Related Work}
\label{sec:related}

\subsection{Neuromorphic Architectures for Spiking Neural Networks}
The development of hardware-compatible SNN architectures has accelerated alongside the maturation of neuromorphic processors such as Intel Loihi, IBM TrueNorth and BrainScaleS. The motivation across this body of work is that SNNs, by virtue of their event-driven binary computation, can in principle replace energy-intensive multiply-accumulate (MAC) operations with sparse synaptic accumulate (AC) operations, achieving substantial reductions in power consumption.

Early convolutional SNN architectures established the feasibility of directly training deep spiking networks via surrogate gradients. The spatio-temporal backpropagation framework of Wu \emph{et al.}~\cite{Wu_2017} unrolled leaky integrate-and-fire dynamics across time and approximated the non-differentiable spike function with a smooth surrogate. Deep residual SNNs followed: SEW-ResNet~\cite{Fang_2021} and MS-ResNet~\cite{Hu_2021} demonstrated that residual connections could be applied within spiking networks, with membrane-shortcut designs guaranteeing strictly binary spike communication and thus efficient hardware deployment.

To bridge the energy and accuracy gap with transformer-based ANNs, Spikformer~\cite{Zhou_2023} introduced spiking self-attention with binary query, key and value tensors. The Spike-driven Transformer~\cite{Yao_2023} pushed this further by redesigning attention as a mask-and-add operation, reducing all components to sparse addition.

Beyond classification, SNNs have been extended to time series~\cite{Lv_2024} and to audio-visual settings. He \emph{et al.}~\cite{He_2025} propose the S-CMRL framework, combining cross-modal complementary attention with semantic-alignment loss. Li \emph{et al.}~\cite{Li_2024} introduce a Tucker-fusion transformer to couple binary spike sequences with floating-point representations. Liu \emph{et al.}~\cite{Liu_2024} take a human-inspired approach using dynamic-vision-sensor lip events as cues for cross-modal attention. These architectures grapple with ensuring the full network operates in a spike-driven manner. The AKD1000 imposes a stricter regime, since it supports only sequential convolutional inference, precluding recurrence and attention. We therefore build on the convolutional SNN literature and treat spike encoder design and firing-rate control as the primary levers for accuracy and energy.

\subsection{Applications of the BrainChip AKD1000}
The AKD1000 is a first-generation digital neuromorphic system-on-chip whose architecture supports the conversion of pre-trained CNNs into SNN-compatible models via the MetaTF/CNN2SNN toolchain. Lunghi \emph{et al.}~\cite{Lunghi_2025} provide the most rigorous quantitative evaluation to date in a space-applications context, reporting EuroSAT inference energies of 0.63--1.38\,mJ per frame at 0.66--1.41\,ms latency for 4-bit-quantized CNNs. They also identify a 911\,mW idle floor that dominates the runtime budget on resource-constrained platforms. Chemnitz and Ermis~\cite{Chemnitz_2025} compare the AKD1000 directly to an NVIDIA GTX 1080 and report 99.5\% lower energy and 76.7\% lower latency on a GXNOR MNIST classifier, with the energy advantage maintained but the latency margin shrinking on a YOLOv2 detector. Lenz and McLelland~\cite{Lenz_2024} apply the AKD1000 to maritime ship detection in satellite imagery via a two-stage AkidaNet/YOLOv5 pipeline, reducing total energy to less than a quarter of a Jetson-Nano-only baseline. Benoot \emph{et al.}~\cite{Benoot_2024} integrate the AKD1000 (and the forthcoming AKD1500) into a heterogeneous on-board satellite data-processing unit. In the biomedical domain, Lutes \emph{et al.}~\cite{Lutes_2024} exploit on-chip edge learning for individualised braking-intent EEG classification, and Br{\aa}tman and Dow~\cite{Bratman_2023} characterise edge-learning hyperparameters for intracranial-pathology CT classification. Across these works, the AKD1000 consistently delivers strong energy advantages on lightweight, sparse inference, while its primary limits are single-node model capacity, the high idle floor and constraints on architectural primitives.

\subsection{AVSR Datasets and Audio-Only SNN Benchmarks}
The GRID corpus~\cite{Cooke_2006} is the standard sentence-level benchmark for lip reading. LRW~\cite{Chung_2016a}, LRS2~\cite{Afouras_2018a}, LRS3~\cite{Afouras_2018}, TCD-TIMIT~\cite{Harte_2015}, AVSpeech~\cite{Ephrat_2018} and ASPIRE~\cite{Gogate_2020} extend coverage to in-the-wild sentences and noisy conditions. On the SNN side, DVS-Lip-Audio~\cite{Tan_2022} provides event-based audio-visual lip data, while SHD/SSC~\cite{Cramer_2019} and the broader Speech Commands corpus~\cite{Warden_2018} provide audio-only benchmarks. Detailed descriptions of the corpora used in this work are deferred to Section~\ref{sec:datasets}.

\subsection{Lip Reading on the GRID Corpus}
\label{ssec:lipreading}
Published evaluation on GRID is almost exclusively visual-only. LipNet~\cite{Assael_2016} established the modern baseline by combining three-dimensional spatio-temporal CNNs with bidirectional GRUs and connectionist temporal classification (CTC) loss, mapping mouth-region frames directly to character sequences and achieving 4.8\% WER on the overlapped split (11.4\% WER unseen-speaker). The Watch, Listen, Attend, and Spell (WLAS) architecture~\cite{Chung_2016} added attention-based sequence-to-sequence decoding and curriculum learning, reaching 3.0\% WER. LCANet~\cite{Xu_2018a} addressed CTC's conditional-independence assumption via cascaded attention-CTC decoding, reaching 2.9\% WER. LipSound~\cite{Qu_2019} reconstructed the mel-spectrogram from lip video and ran ASR on the result, reaching 2.5\% WER. DualLip~\cite{Chen_2020b} introduced a generation/recognition dual learning scheme. HLR-Net~\cite{Sarhan_2021} combined inception modules with attention-CTC, reaching 3.3\% WER overlapped and 9.7\% WER unseen-speaker. LCSNet~\cite{Xue_2022} added channel-attention and selective-feature fusion, reaching 2.3\% WER. Most recently, the landmark-guided cross-speaker model of Wu \emph{et al.}~\cite{Wu_2024}, built on a hybrid CTC/attention conformer back-end with mutual-information regularisation, reached 1.83\% WER overlapped and 10.21\% WER on the unseen-speaker split. These results are summarised in Table~\ref{tab:grid_sota}.

\begin{table}[t]
\centering
\caption{State-of-the-art lip-reading WER on the GRID corpus. All baselines use full 3D convolutions, recurrent layers or attention, none of which are supported on AKD1000.}
\label{tab:grid_sota}
\begin{tabular}{@{}lccc@{}}
\toprule
Model & Year & Setting & WER (\%) \\
\midrule
LipNet~\cite{Assael_2016}            & 2016 & Overlapped & 4.80 \\
WLAS~\cite{Chung_2016}                & 2017 & Overlapped & 3.00 \\
LCANet~\cite{Xu_2018a}                & 2018 & Overlapped & 2.90 \\
LipSound~\cite{Qu_2019}               & 2019 & Overlapped & 2.50 \\
DualLip~\cite{Chen_2020b}             & 2020 & Overlapped & 2.71 \\
HLR-Net~\cite{Sarhan_2021}            & 2021 & Overlapped & 3.30 \\
LCSNet~\cite{Xue_2022}                & 2023 & Overlapped & 2.30 \\
Wu \emph{et al.}~\cite{Wu_2024}       & 2024 & Overlapped & 1.83 \\
\midrule
LipNet~\cite{Assael_2016}            & 2016 & Unseen     & 11.40 \\
HLR-Net~\cite{Sarhan_2021}            & 2021 & Unseen     & 9.70 \\
Wu \emph{et al.}~\cite{Wu_2024}       & 2024 & Unseen     & 10.21 \\
\bottomrule
\end{tabular}
\end{table}

Several trends are noteworthy. First, attention-augmented decoders consistently improve over CTC alone by better modelling output dependencies. Second, the visual front-end remains a primary bottleneck, and methods that improve it through channel attention, landmark localisation or intermediate acoustic reconstruction yield large gains. Third, the gap between overlapped and unseen-speaker performance remains substantial. Finally, no published results exist for full AVSR fusion on GRID under noisy audio conditions, presenting both a gap in the literature and a relevant comparison point for any neuromorphic approach.

\section{Proposed System}
\label{sec:method}

\subsection{Architecture Overview}
\begin{figure*}[t]
\centering
\includegraphics[width=0.95\textwidth]{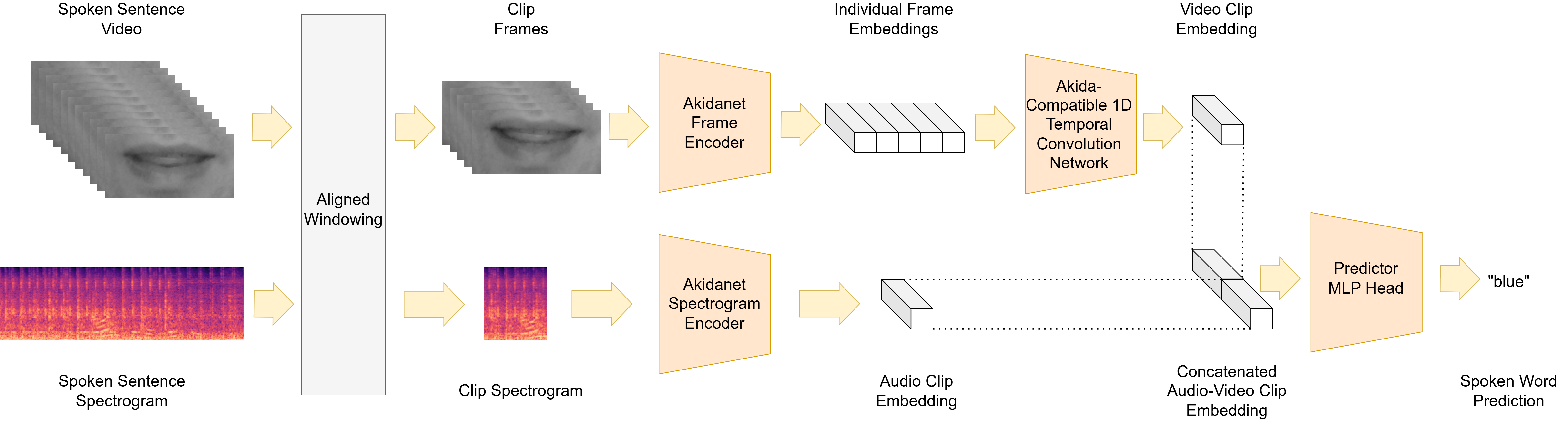}
\caption{NAVIR audio-visual pipeline. Aligned video and audio windows are independently encoded by AkidaNet-based per-frame, temporal-video and spectrogram-audio encoders. Embeddings are concatenated and fed to an MLP predictor head, whose CTC output is decoded by constrained beam search. All blocks are AKD1000-compatible (no 3D convolutions, recurrence or attention).}
\label{fig:pipeline}
\end{figure*}

The system takes as input the cropped lip region of a speaker, extracted from video using the off-the-shelf MediaPipe Face Mesh model, alongside the corresponding audio signal. From these two streams, the model produces spoken-word predictions at the clip level through a multi-stage audio-visual pipeline shown in Figure~\ref{fig:pipeline}.

\subsection{Aligned Windowing}
\label{ssec:windowing}
Raw video frames and audio are first segmented into overlapping clips using a sliding window. The video window is parameterised by its size in frames and a step size, set to roughly one third of the window size so adjacent windows share substantial overlap. This avoids missed words at boundaries and produces a dense sequence of clip-level predictions that the decoder later aggregates.

The audio (spectrogram) window is treated as an independent hyperparameter. It is expressed in spectrogram time bins and is in general not equal to the video window size, since the two streams are sampled at different rates and the optimal acoustic context may differ from the optimal visual context. To keep the modalities synchronised at the predictor input, every audio window is centre-aligned with its corresponding video window. Concretely, given a video window centred at time $t_c$, the audio window is taken as the spectrogram segment of the configured length whose centre also lies at $t_c$, regardless of the relative widths of the two windows. This decoupling lets the audio context be widened or narrowed without changing the video tiling, while preserving a one-to-one temporal correspondence between video and audio embeddings at the fusion stage.

\subsection{Frame-Level Visual Encoding}
Each frame in a clip is independently encoded into a compact embedding using an AkidaNet-based CNN, an architecture purpose-designed for the BrainChip Akida hardware. Processing frames individually rather than as a volumetric sequence is a deliberate architectural choice, since the AKD1000 does not support 3D convolutions, recurrent connections or attention. Factorising encoding into a spatial stage (per frame) and a temporal stage (across frames) keeps the entire pipeline hardware-compatible.

\subsection{Video-Clip Temporal Encoding}
Once each frame has been encoded, the resulting embeddings are stacked along the time axis into a 2D representation of the clip (time $\times$ embedding features). A second Akida-compatible network applies temporal convolutions across this stack, producing a single video-clip embedding that captures motion and short-range temporal dynamics.

\subsection{Spectrogram-Clip Audio Encoding}
In parallel, the audio corresponding to each clip is converted into Mel-frequency cepstral coefficients (MFCCs), the standard CNN-friendly audio representation. The spectrogram is processed by a third AkidaNet-based encoder to yield an audio-clip embedding.

\subsection{Predictor Head and Audio-Visual Fusion}
The video-clip embedding and the audio-clip embedding are concatenated into a joint audio-visual representation, fed to a lightweight MLP predictor head that outputs a vector of logits over the vocabulary. Per-token probabilities are obtained via softmax.

\subsection{Constrained Beam-Search Decoding}
At inference time, the head produces a token-score matrix of shape $T \times V$ (clip windows $\times$ vocabulary). Rather than greedy decoding, we use a constrained beam search exploiting the known grammar of the evaluation datasets. Both corpora consist of sentences drawn from a fixed, finite grammar. The decoder restricts every per-frame transition to tokens that form a valid prefix of some legal sentence, dramatically shrinking the search space.

The beam search maintains the top-$B$ active hypotheses, each represented as a partial token prefix and an accumulated score. At each frame, three transition types are considered: emitting a blank or silence token (prefix unchanged), repeating the last emitted token (absorbing duplicate frame predictions), or advancing to a new token drawn from the legal-next-token set for the current prefix. Beams are pruned to the top $B$ after expansion, and the best fully terminated hypothesis is tracked throughout. If no beam reaches a valid terminal sentence, a fallback extends the best partial prefix into the nearest valid sentence. If no extension exists, the prefix is progressively shortened, ultimately defaulting to the globally shortest valid sentence. This guarantees grammatically valid output at all times.

\subsection{Training and Quantization}
\label{ssec:training}
The model is trained with CTC loss, which marginalises over all alignments between input frames and target tokens via a special blank label, allowing weakly supervised training where only the spoken sentence is annotated. Word error rate (WER), the minimum edit distance between prediction and reference normalised by reference length, is the primary metric. For the industrial-command corpus we additionally report sentence-level command accuracy.

The Akida hardware requires integer-precision weights and activations. Quantization is parameterised by three bit-widths, written as a triple $w_{\text{in}}/w/a$, where $w_{\text{in}}$ is the weight bit-width of the first layer, $w$ the weight bit-width of all subsequent layers and $a$ the activation bit-width. Standard AKD1000-compatible models use 8/4/4 throughout. Our modular pipeline uses a hybrid scheme. The image encoder and the spectrogram encoder, which receive raw pixel and spectrogram inputs respectively, use 8/4/4. The video encoder and the predictor head, which operate on intermediate embeddings produced upstream, use 4/4/4. Quantization is followed by quantization-aware training (QAT) to recover any post-quantization accuracy loss. Throughout both float and QAT phases we use only $\ell_2$ weight decay (the same regulariser as the original AkidaNet recipe~\cite{Lunghi_2025}). No magnitude pruning or $\ell_1$ penalty is applied.

\section{Datasets}
\label{sec:datasets}

\subsection{GRID Audiovisual Sentence Corpus}
GRID~\cite{Cooke_2006} is the canonical sentence-level lip-reading benchmark and our primary public evaluation. It contains audio-visual recordings from 34 speakers, each uttering 1{,}000 sentences drawn from a fixed six-word grammar:
\begin{quote}
\textit{command (4) + colour (4) + preposition (4) + letter (25) + digit (10) + adverb (4)}
\end{quote}
For example, ``set blue with H seven again''. The grammar yields up to 64{,}000 distinct sentences. All clips are approximately 3\,s long, recorded at 25\,fps and 720$\times$576 pixels, and ship with word-level alignments. Standard protocols use either an overlapped split (some speakers in both train and test sets) or an unseen-speaker split. We chose GRID for its public availability, manageable size and the abundance of published baselines (Section~\ref{ssec:lipreading}).

\subsection{NAVIR Industrial-Command Corpus (Internal)}
\label{ssec:navir_dataset}
The NAVIR corpus is a custom audio-visual dataset of robot-manipulation commands developed for this project. It is an \emph{internal corpus}, used to fine-tune and evaluate the deployed system but not publicly released. The corpus consists of 183 distinct commands drawn from a structured vocabulary of approximately 39 words, organised in five categories.
\begin{enumerate}
\item \emph{Move [object] into [location]} (e.g., ``Move the blue cube into the box'').
\item \emph{Pick up [object]} (e.g., ``Grab the mouse'').
\item \emph{Place [object] in [location]} (e.g., ``Put the green cube in the box'').
\item \emph{Go to [object/location/position]} (e.g., ``Go above the blue cube'', ``Return to home'').
\item \emph{Rotate [object] clockwise} (e.g., ``Spin the battery clockwise'').
\end{enumerate}
Objects are: blue cube, yellow cube, green cube, white ball, mouse, battery. Locations are: box, bin. Positional targets are: home, up, ready. Each command admits multiple synonymous phrasings (e.g., \emph{move / transfer / relocate / shift / bring}), introducing lexical variation while preserving semantic equivalence. Two speakers each produced the full set of 183 commands, yielding 366 recordings. Compared to GRID, NAVIR is domain-specific and command-oriented, making it more representative of the target deployment.

\subsection{UrbanSound8K Noise Augmentation}
UrbanSound8K~\cite{Salamon_2014} is a publicly available corpus of 8{,}732 labelled urban-sound excerpts of up to 4\,s, drawn from ten classes (air conditioner, car horn, children playing, dog bark, drilling, engine idling, gun shot, jackhammer, siren and street music) and pre-folded for ten-fold cross-validation. We use it solely as a noise-augmentation source. Following CochleaNet~\cite{Gogate_2020a}, we extract the mechanical and machinery subset (air conditioner, drilling, engine idling, jackhammer), which is representative of the ambient noise found in industrial settings, and mix these clips into the clean audio tracks of GRID and the NAVIR corpus at SNRs sampled from $\{-15, -10, -5, 0\}$\,dB during training. For evaluation, the noisy-audio test condition uses a fixed SNR of $-10$\,dB. The first predefined fold is held out for evaluation.

\section{Experimental Setup}
\label{sec:setup}

\begin{table}[t]
\centering
\caption{Per-dataset training and pre-processing hyperparameters.}
\label{tab:setup}
\setlength{\tabcolsep}{3pt}
\footnotesize
\begin{tabular}{@{}lcc@{}}
\toprule
& GRID & NAVIR \\
\midrule
Batch size                   & 16        & 3 \\
Epochs, float (unseen/overlap) & 30 / 100  & --- / 200 \\
Epochs, QAT (unseen/overlap)   & 15 / 100  & --- / 200 \\
Video window (frames)        & 15        & 18 \\
Video overlap (frames)       & 10        & 12 \\
Audio (spectrogram) window   & 60        & 60 \\
Sample rate (Hz)             & 50{,}000  & 32{,}000 \\
FFT points                   & 2{,}048   & 1{,}024 \\
Hop length                   & 512       & 320 \\
Mel bands                    & 112       & 112 \\
Noise augmentation ratio     & 0.8       & 0.8 \\
SNR sweep (dB)               & $\{-15,-10,-5,0\}$ & $\{-15,-10,-5,0\}$ \\
Image encoder ($\alpha$)     & 0.50      & 0.25 \\
Spec encoder ($\alpha$)      & 0.50      & 0.25 \\
Video encoder dim ($\alpha$) & 256 (1.0) & 128 (0.5) \\
Predictor head (units)       & 512, 256  & 256 \\
\bottomrule
\end{tabular}
\end{table}

Table~\ref{tab:setup} summarises the configuration of both training runs. On GRID we report both the standard unseen-speaker protocol of LipNet~\cite{Assael_2016} and an overlapped-speaker protocol (speakers shared between train and test, sentences disjoint), which is the other widely reported GRID setting. NAVIR uses an 80/20 sentence-level split that holds out a random subset of commands across both speakers, so the evaluation set probes generalisation to unseen \emph{sentences} rather than unseen speakers (the corpus contains only two speakers). Both runs apply horizontal-flip augmentation ($p=0.5$) and temporal jitter ($p=0.05$). After float training, models are quantized using the hybrid scheme of Section~\ref{ssec:training} and fine-tuned with QAT (15 epochs on GRID unseen-speaker, 100 epochs on GRID overlapped, 200 epochs on NAVIR; the longer NAVIR schedule reflects the smaller corpus and the longer overlapped schedule reflects the harder cross-sentence generalisation within a fixed speaker set).

\section{Recognition Results}
\label{sec:results}

\subsection{GRID Word Error Rate}
Tables~\ref{tab:grid_float} and~\ref{tab:grid_qat} report WER under the standard unseen-speaker split, and Tables~\ref{tab:grid_float_overlap} and~\ref{tab:grid_qat_overlap} report WER under the overlapped-speaker split. Results are broken down by training modality (clean audio, noisy audio, video only, or fused) and evaluation condition (clean / noisy audio).

\begin{table}[t]
\centering
\caption{Non-quantized WER (\%) on GRID (unseen-speaker split) under clean and noisy audio test conditions.}
\label{tab:grid_float}
\begin{tabular}{@{}lcc@{}}
\toprule
Training modalities    & Clean audio & Noisy audio \\
\midrule
Clean audio            & 3.7   & 79.9 \\
Noisy audio            & 4.4   & 21.0 \\
Video                  & 34.0  & 34.0 \\
Noisy audio + video    & 7.7   & 16.6 \\
Clean audio + video    & 3.6   & 78.6 \\
\bottomrule
\end{tabular}
\end{table}

\begin{table}[t]
\centering
\caption{Quantized WER (\%) on GRID (unseen-speaker split) after QAT.}
\label{tab:grid_qat}
\begin{tabular}{@{}lcc@{}}
\toprule
Training modalities    & Clean audio & Noisy audio \\
\midrule
Clean audio            & 4.2   & 77.3 \\
Noisy audio            & 5.2   & 22.5 \\
Video                  & 35.3  & 35.3 \\
Noisy audio + video    & 5.3   & 14.0 \\
Clean audio + video    & 3.2   & 77.8 \\
\bottomrule
\end{tabular}
\end{table}

\begin{table}[t]
\centering
\caption{Non-quantized WER (\%) on GRID (overlapped-speaker split) under clean and noisy audio test conditions. Both float and QAT models were trained for 100 epochs in this regime.}
\label{tab:grid_float_overlap}
\begin{tabular}{@{}lcc@{}}
\toprule
Training modalities    & Clean audio & Noisy audio \\
\midrule
Clean audio            & 1.3   & 79.8 \\
Noisy audio            & 1.9   & 10.4 \\
Video                  & 9.1   & 9.1 \\
Noisy audio + video    & 0.7   & 2.8 \\
Clean audio + video    & 0.7   & 77.4 \\
\bottomrule
\end{tabular}
\end{table}

\begin{table}[t]
\centering
\caption{Quantized WER (\%) on GRID (overlapped-speaker split) after QAT.}
\label{tab:grid_qat_overlap}
\begin{tabular}{@{}lcc@{}}
\toprule
Training modalities    & Clean audio & Noisy audio \\
\midrule
Clean audio            & 2.0   & 78.7 \\
Noisy audio            & 2.3   & 11.8 \\
Video                  & 6.7   & 6.7 \\
Noisy audio + video    & 0.8   & 3.3 \\
Clean audio + video    & 0.8   & 77.1 \\
\bottomrule
\end{tabular}
\end{table}

Under clean audio, all audio-capable models on the unseen-speaker split perform comparably (3.2--7.7\% WER across configurations). The interesting contrast appears under noise. Training on clean audio collapses to roughly 77--80\% WER when noise is introduced, while training on noisy audio brings this down to 21.0\% (float) and 22.5\% (quantized). Crucially, fusing noisy audio with video reduces WER further to 16.6\% (float) and 14.0\% (quantized), confirming that the visual modality anchors recognition when the acoustic signal degrades. Interestingly, in this fused noisy condition the quantized model is marginally better than its float counterpart, although the effect is small and not consistent across all configurations.

The video-only model reaches 34.0\% (float) and 35.3\% (quantized) WER on the unseen-speaker split. While this is higher than the SOTA range of 9.7--11.4\% (Section~\ref{ssec:lipreading}), those baselines all employ 3D convolutions, attention-augmented decoders and large-scale pre-training, none of which are AKD1000-compatible. The factorised AkidaNet pipeline trades representational depth for hardware deployability, and even so fusion still extracts useful information from the visual stream when audio is corrupted.

On the overlapped-speaker split (Tables~\ref{tab:grid_float_overlap} and~\ref{tab:grid_qat_overlap}), where train and test sets share speakers but not sentences, every trend observed in the unseen-speaker regime is preserved and the absolute numbers tighten substantially. The video-only model reaches 9.1\% (float) and 6.7\% (quantized) WER, approaching the 1.83--4.8\% range of unconstrained ANN baselines despite the AKD1000-imposed architectural budget. Notably, QAT improves the video-only model over its float counterpart by 2.4 absolute points in this setting, an effect of similar direction to the one observed in the unseen-speaker fused condition, though again we do not claim a consistent QAT advantage across configurations. As in the unseen-speaker case, fusing the two modalities yields a clear improvement over either modality alone: under noise, the noisy-audio + video model reaches 2.8\% (float) and 3.3\% (quantized) WER, against 10.4\% (float) and 11.8\% (quantized) for noisy audio alone, and 9.1\% (float) and 6.7\% (quantized) for video alone, confirming that the visual modality anchors recognition when the acoustic signal degrades. Under clean audio, fusion reduces WER to 0.7\% (float) and 0.8\% (quantized), below both single-modality baselines. As in the unseen-speaker case, training on clean audio alone collapses under noise (77.1--79.8\% WER across the four clean-audio-trained configurations), whereas training on noisy audio reduces noisy-test WER by nearly an order of magnitude (10.4\% float and 11.8\% quantized). The Pareto analysis in Section~\ref{ssec:pareto} uses both the overlapped- and unseen-speaker video-only stats for cross-method comparison.

\subsection{NAVIR Word Error Rate and Command Accuracy}
Tables~\ref{tab:navir_float} and~\ref{tab:navir_qat} report WER and sentence-level command accuracy on the NAVIR corpus.

\begin{table}[t]
\centering
\caption{Non-quantized NAVIR results: WER (\%) / command accuracy (\%).}
\label{tab:navir_float}
\begin{tabular}{@{}lcc@{}}
\toprule
Training modalities & Clean audio & Noisy audio \\
\midrule
Clean audio            & 5.4 / 93.0   & 97.5 / \phantom{0}2.8 \\
Noisy audio            & 11.0 / 81.7  & 42.6 / 52.1 \\
Video                  & 0.5 / 100.0  & 0.5 / 100.0 \\
Noisy audio + video    & 0.3 / 100.0  & 0.9 / 100.0 \\
Clean audio + video    & 4.3 / 93.0   & 95.6 / \phantom{0}2.8 \\
\bottomrule
\end{tabular}
\end{table}

\begin{table}[t]
\centering
\caption{Quantized NAVIR results after QAT: WER (\%) / command accuracy (\%).}
\label{tab:navir_qat}
\begin{tabular}{@{}lcc@{}}
\toprule
Training modalities & Clean audio & Noisy audio \\
\midrule
Clean audio            & 6.6 / 91.5   & 98.7 / \phantom{0}0.0 \\
Noisy audio            & 12.1 / 77.5  & 43.1 / 47.9 \\
Video                  & 0.7 / 100.0  & 0.7 / 100.0 \\
Noisy audio + video    & 0.6 / 98.6   & 1.5 / 98.6 \\
Clean audio + video    & 4.3 / 93.0   & 95.7 / \phantom{0}4.2 \\
\bottomrule
\end{tabular}
\end{table}

NAVIR results are strong across vision-based modalities both before and after quantization. The video-only model retains near-perfect performance through quantization (0.5\% to 0.7\% WER, 100\% command accuracy preserved), and the noisy-audio + video fusion model degrades only marginally (0.3\% to 0.6\% WER clean and 0.9\% to 1.5\% noisy). Audio-only models are far less robust. The clean-audio model collapses entirely under noise both before (97.5\%) and after (98.7\%) quantization, confirming that without visual input the system has no path to reliable performance in real-world acoustic environments. The 200-epoch QAT schedule is well matched to the small corpus size, with multimodal models retaining most of their float accuracy.

\subsection{Cross-Dataset Discussion}
Across both benchmarks, multimodal fusion provides a robust gain over single-modality baselines, especially under noise. QAT preserves accuracy and in some configurations slightly improves it. The video-only ceiling on the GRID unseen-speaker split (35.3\% WER) reflects the architectural constraints of the AKD1000 rather than a fundamental limit of lip reading itself. On NAVIR, where the visual task is much easier (small command vocabulary, controlled recording conditions, two speakers), the same architecture saturates near 0\% WER.

\section{Energy and Power Analysis}
\label{sec:energy}

\subsection{Theoretical Energy Methodology}
We adopt the operation-count framework that has become a standard in the SNN/ANN comparison literature~\cite{Zhou_2023,Hu_2021,Liu_2024,Kundu_2021}, parameterised by the per-operation energy constants reported by Horowitz~\cite{Horowitz_2014} for 45\,nm CMOS. Horowitz~\cite{Horowitz_2014} does not propose a model-level methodology, only the underlying primitive-cost numbers. The methodology consists of summing the relevant operations in the model and weighting them by these primitive costs. The energy of a conventional ANN inference is
\begin{equation}
E_\mathrm{ANN} = \mathrm{MACs} \times E_\mathrm{MAC},
\end{equation}
with $E_\mathrm{MAC} \approx 3.7$\,pJ. For an SNN, synaptic operations reduce to additions, since spike values are binary and inactive neurons contribute nothing, at $E_\mathrm{AC} \approx 0.9$\,pJ, scaled by the empirically measured average firing rate $\bar r$:
\begin{equation}
E_\mathrm{SNN} = \mathrm{MACs} \times \bar r \times E_\mathrm{AC}.
\end{equation}
The combined effect of spiking sparsity and the lower per-operation cost yields a theoretical efficiency gain
\begin{equation}
\frac{E_\mathrm{ANN}}{E_\mathrm{SNN}} = \frac{E_\mathrm{MAC}}{E_\mathrm{AC} \times \bar r} = \frac{3.7}{0.9 \times \bar r}.
\end{equation}
These figures are theoretical estimates derived from operation counts. They do not capture memory-access cost, hardware parallelism or implementation-specific factors, but they serve as a standard, reproducible cross-architecture baseline.

\subsection{Theoretical Results on GRID}
Table~\ref{tab:complexity} reports parameter counts, FLOPs and ANN inference energy for our video-only model alongside two SOTA lip-reading baselines, accounting for the total number of model calls required for a sentence-level prediction. Our model is substantially more compact, with 3.1$\times$ fewer FLOPs than LipNet~\cite{Assael_2016} and 37.8$\times$ fewer than Wu \emph{et al.}~\cite{Wu_2024}. This translates directly into a 4.8$\times$ and 37.8$\times$ lower ANN energy footprint respectively.

\begin{table}[t]
\centering
\caption{Model complexity and theoretical ANN inference energy on GRID.}
\label{tab:complexity}
\begin{tabular}{@{}lcccc@{}}
\toprule
Model & WER (\%) & Params & FLOPs & ANN energy \\
\midrule
LipNet~\cite{Assael_2016}      & 11.4  & 4.57\,M  & 10.69\,G  & 19.78\,mJ \\
Wu \emph{et al.}~\cite{Wu_2024} & 10.21 & 10.89\,M & 84.62\,G  & 156.54\,mJ \\
Ours (video-only)              & 35.30 & 1.49\,M  & 2.24\,G   & 4.15\,mJ \\
\bottomrule
\end{tabular}
\end{table}

The SNN energy advantage stems from the combination of spike sparsity and the AC-vs-MAC asymmetry. Our model achieves a measured average firing rate of 27.55\% (i.e., 72.45\% sparsity) across spiking layers, evaluated empirically on GRID. Substituting this into the SNN formula yields a theoretical 13.17$\times$ gain over the equivalent ANN, reducing per-sentence inference cost to 314.92\,$\mu$J. Combined with the already-low ANN baseline of 4.15\,mJ, this places our model approximately 62.8$\times$ below LipNet~\cite{Assael_2016} and 497$\times$ below Wu \emph{et al.}~\cite{Wu_2024} when both are evaluated as ANNs (Table~\ref{tab:snn_energy}). LipNet and Wu \emph{et al.} are not directly convertible to SNN equivalents and are quoted as ANN comparison points only.

\begin{table}[t]
\centering
\caption{Theoretical SNN energy and efficiency gain.}
\label{tab:snn_energy}
\begin{tabular}{@{}lccc@{}}
\toprule
Model & ANN energy & SNN energy & ANN/SNN gain \\
\midrule
LipNet~\cite{Assael_2016}        & 19.78\,mJ   & ---          & --- \\
Wu \emph{et al.}~\cite{Wu_2024}  & 156.54\,mJ  & ---          & --- \\
Ours (video-only)                & 4.15\,mJ    & 314.92\,$\mu$J & 13.17$\times$ \\
\bottomrule
\end{tabular}
\end{table}

\subsection{Pareto Analysis of Accuracy vs.\ Computational Cost}
\label{ssec:pareto}

\begin{figure}[t]
\centering
\includegraphics[width=0.95\columnwidth]{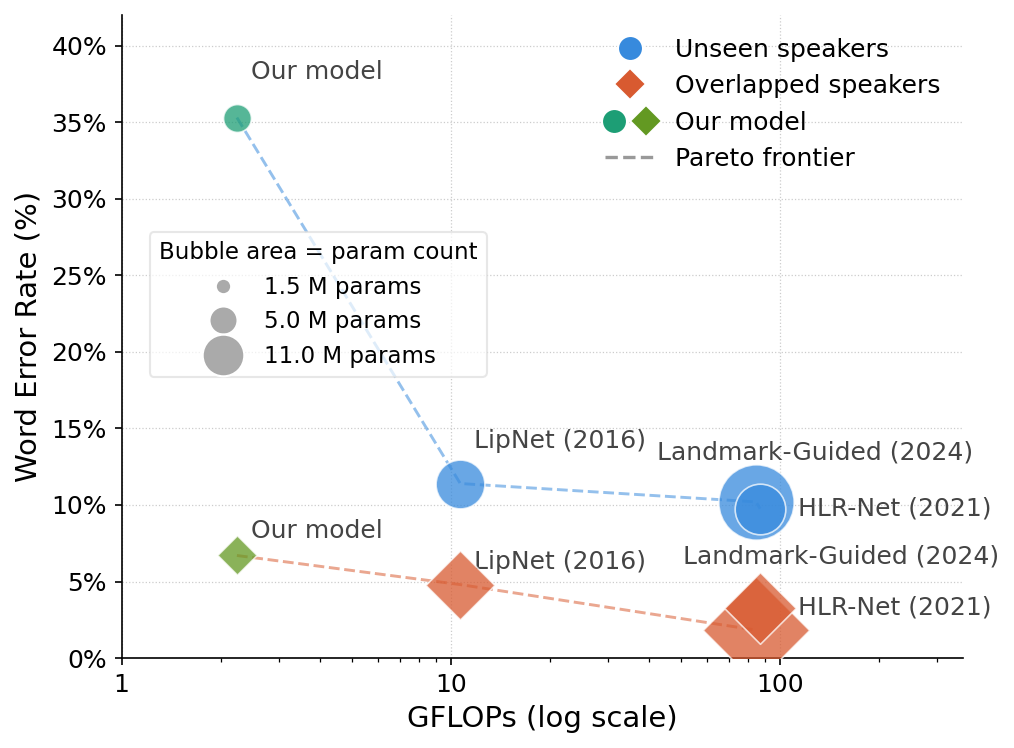}
\caption{Pareto view of GRID lip-reading models, computational cost (GFLOPs, log scale) versus word error rate, for both overlapped-speaker (red diamonds) and unseen-speaker (blue circles) test conditions. Bubble area is proportional to parameter count. Our quantized video-only model (green) sits alone on the low-FLOPs frontier.}
\label{fig:pareto}
\end{figure}

Figure~\ref{fig:pareto} situates our model in the GFLOPs--WER plane against the SOTA lip readers reviewed in Section~\ref{ssec:lipreading}. Two test conditions are plotted: unseen-speaker (blue circles) and overlapped-speaker (red diamonds). Bubble area encodes parameter count. Our model is the only point at low FLOPs (2.24\,GFLOPs) and the smallest by parameters (1.49\,M), while the SOTA cluster (LipNet, HLR-Net, Wu \emph{et al.}) sits at 10--85\,GFLOPs and 4--11\,M parameters, achieving 1.83--11.4\% WER. We are deliberately Pareto-incomparable. At the cost of higher unseen-speaker WER, NAVIR offers an order of magnitude or more lower computational cost, which is the relevant axis for neuromorphic edge deployment. We also note that the overlapped-speaker number for our quantized video-only model (6.7\% WER) is the first reported result for an SNN-compatible architecture on GRID and approaches the 1.83--4.8\% range of unconstrained ANN baselines despite our fixed-quantization, no-3D-conv, no-attention budget.

\subsection{Practical Hardware Power Measurements}
To quantify the AKD1000's practical efficiency, we measured power consumption on the Raspberry Pi 5 demo platform with a FNIRSI FNB58 USB power meter, repeatedly calling the model on pre-computed inputs over a 5-minute window so that average draw stabilised. We also ran the same benchmark on a laptop with an NVIDIA RTX 3060 Mobile GPU using continuous \texttt{nvidia-smi} polling. The two measurement scopes differ: the Pi figures cover total system draw, whereas the GPU figures cover only the GPU itself, with known precision limitations~\cite{Yang_2024}. Both effects make the GPU comparison favourable to the GPU.

The CPU-backend idle draw on the Pi settled at approximately 390\,mWh / 5\,min, which we adopt as the baseline for inference-attributable consumption. The Akida-backend idle draw is essentially identical (387.59\,mWh), and the bare Pi with no script running drew 370.92\,mWh. The GPU's idle draw alone was 1{,}280\,mWh / 5\,min, more than three times the Pi's total system draw.

\begin{table}[t]
\centering
\caption{Video-only model: per-inference power consumption across backends. Idle baselines: Pi 390\,mWh / 5\,min, GPU 1{,}280\,mWh / 5\,min. Pi figures cover total system draw (FNB58). GPU figures cover GPU-only draw via \texttt{nvidia-smi}, normalised to a 5-minute window.}
\label{tab:power_video}
\begin{tabular}{@{}lcccc@{}}
\toprule
Backend & it/s & mWh@5\,min & $\Delta$mWh & mWh/inf. \\
\midrule
Pi + AKD1000 (Akida)  & 14.55 & \phantom{0}462.0 & \phantom{0}72.0  & 0.0165 \\
Pi + CPU (Akida)      & 15.55 & \phantom{0}768.0 & 378.0            & 0.0810 \\
Pi + CPU (Keras)      & 1.10  & \phantom{0}531.9 & 141.9            & 0.4306 \\
Laptop + GPU (Keras)  & 2.54  & 2{,}569.6        & 1{,}289.6        & 1.6913 \\
\bottomrule
\end{tabular}
\end{table}

\begin{table}[t]
\centering
\caption{Audio-video model: per-inference power consumption across backends. Same idle baselines and measurement scopes as Table~\ref{tab:power_video}.}
\label{tab:power_av}
\begin{tabular}{@{}lcccc@{}}
\toprule
Backend & it/s & mWh@5\,min & $\Delta$mWh & mWh/inf. \\
\midrule
Pi + AKD1000 (Akida)  & 2.61  & \phantom{0}459.9 & \phantom{0}69.9  & 0.0894 \\
Pi + CPU (Akida)      & 13.36 & \phantom{0}737.2 & 347.2            & 0.0866 \\
Pi + CPU (Keras)      & 0.63  & \phantom{0}528.5 & 138.5            & 0.7381 \\
Laptop + GPU (Keras)  & 1.29  & 2{,}566.2        & 1{,}286.2        & 3.3354 \\
\bottomrule
\end{tabular}
\end{table}

The most informative on-device comparison is the AKD1000 against the same SNN-converted model running on CPU under the Akida software backend (the strongest CPU baseline). The Pi-CPU Keras backend, which executes the float Keras model rather than the SNN, is reported alongside for completeness but is the weakest competitor. Comparing AKD1000 against laptop-GPU Keras measures the chip's standing against a typical accelerated-computing baseline.

The video-only pipeline (Table~\ref{tab:power_video}) consists of an image encoder (75\,NPs, 1 sequence), a temporal encoder (41\,NPs, 3 sequences) and a predictor head (2\,NPs, 1 sequence), for a total of 5 AKD1000 passes per inference. The Akida backend reaches 14.55\,it/s at 0.0165\,mWh/inf. Against the same model on CPU under the Akida backend (0.0810\,mWh/inf., 15.55\,it/s) the AKD1000 is 4.9$\times$ more energy-efficient per inference at comparable throughput. Against the Pi-CPU Keras float baseline the gap widens to 26$\times$. Against the laptop GPU it widens further to over 100$\times$.

The audio-video pipeline (Table~\ref{tab:power_av}) extends the video-only pipeline with a spectrogram encoder (348\,NPs, 9 sequences) and a larger predictor head (2\,NPs, 1 sequence), for a total of 22 passes per inference. Per-inference Akida energy rises to 0.0894\,mWh, broadly comparable to the Pi-CPU Akida backend (0.0866\,mWh/inf.) but at lower throughput (2.61 vs.\ 13.36\,it/s). Against the laptop GPU the AKD1000 still consumes 37$\times$ less energy per inference. The reason the AKD1000 loses its per-inference advantage on the audio-video model is a hardware-mapping behaviour of the Akida toolchain that is independent of architecture, analysed in Appendix~\ref{app:av_slower}. Even so, 2.61\,it/s is comfortably above real-time for command recognition.

\section{Demonstration System}
\label{sec:demo}

\begin{figure}[t]
\centering
\includegraphics[width=0.95\columnwidth]{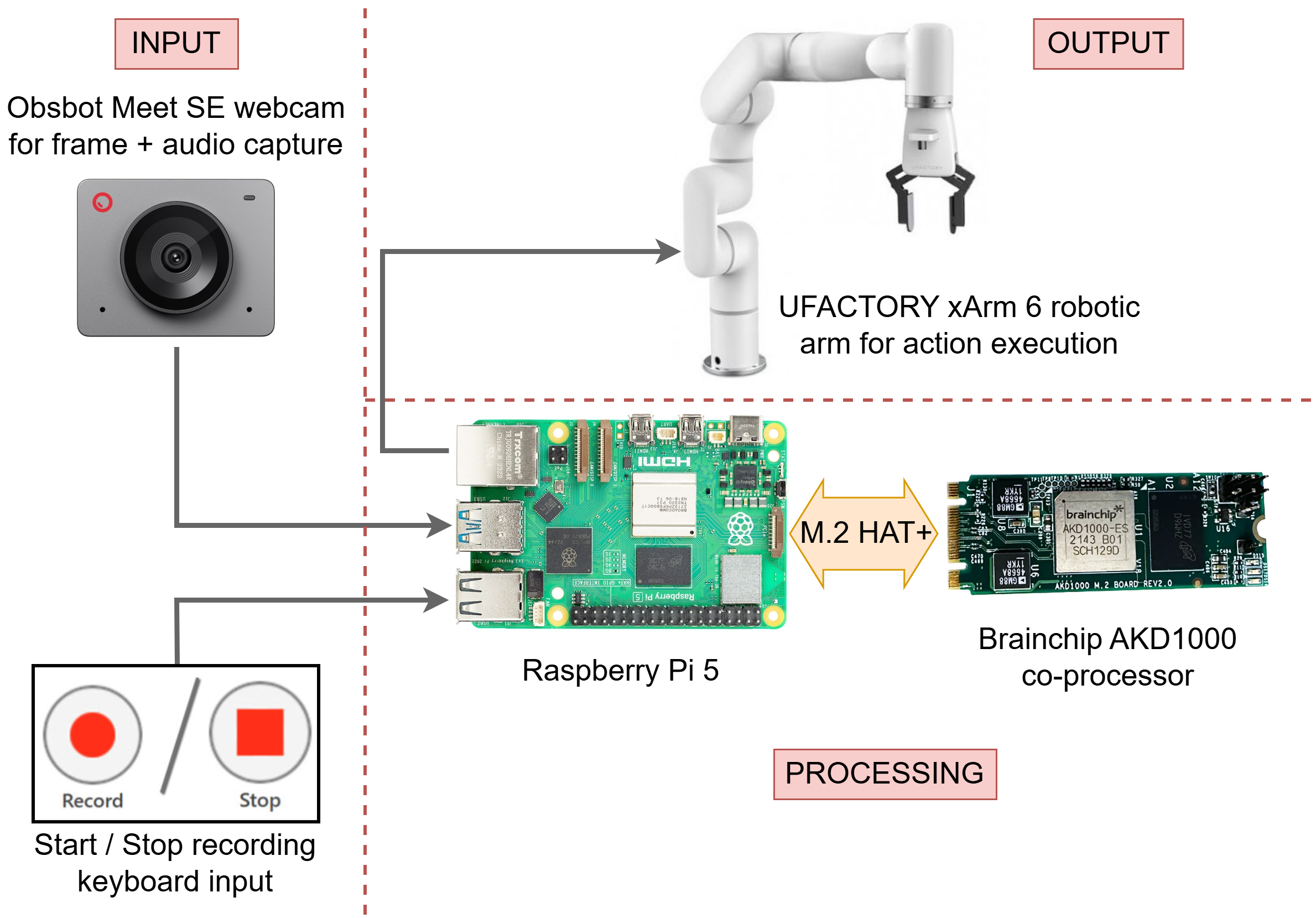}
\caption{NAVIR demonstration system. An Obsbot Meet SE webcam supplies frame and audio capture. A Raspberry Pi 5 with an M.2 HAT hosts the BrainChip AKD1000 co-processor. A uFactory xArm 6 robotic arm executes the recognised commands.}
\label{fig:demo_system}
\end{figure}

The demo system, shown in Figure~\ref{fig:demo_system}, is built around a 16\,GB Raspberry Pi 5 chosen for its balance of processing capability, compact form factor and peripheral-stack compatibility. An M.2 HAT exposes a PCIe-backed M.2 slot through which the AKD1000 M.2 card is connected directly to the host, enabling low-latency communication between the Pi and the neuromorphic accelerator without an external workstation. Visual input comes from an Obsbot Meet SE webcam at full HD, the same model used during NAVIR-corpus collection, ensuring identical lens optics, colour response and field of view at deployment time. Recording start and stop is driven by a standard USB keyboard. A uFactory xArm 6 robotic arm is connected directly to the Pi, closing the loop from voice command to physical action. Video and audio are captured, processed through the neuromorphic pipeline, and the recognised commands are dispatched to the arm in real time. The complete system is self-contained, edge-deployable and runs entirely on commodity embedded hardware augmented by the AKD1000.

\section{Discussion and Limitations}
\label{sec:discussion}

The system has several known limitations. First, the AKD1000's architectural constraints preclude 3D convolutions, recurrence and attention, which together account for most of the SOTA accuracy on unconstrained lip-reading benchmarks. The 35.3\% video-only WER on the GRID unseen-speaker split reflects this gap rather than any limit of the proposed factorisation. Second, the audio-video model's hardware mapping is not yet optimal, as Appendix~\ref{app:av_slower} makes precise. The image encoder spills across 9 sequences in the audio-video checkpoint despite mapping to a single sequence in the video-only checkpoint. Third, the NAVIR corpus, while well matched to the target deployment, contains only 366 recordings from two speakers and does not yet probe unseen-speaker generalisation or broader vocabulary.

Several directions naturally extend this work. The forthcoming AKD1500, which promises lower static power and supports more complex topologies, may relax some current constraints. Sparsity-aware fine-tuning, motivated by the mapping analysis in Appendix~\ref{app:av_slower}, could equalise the audio-video and video-only sequence counts and yield a one-shot throughput gain on existing hardware. Expanding the NAVIR corpus along the speaker, vocabulary and noise axes would strengthen the system's claim to industrial readiness. The AKD1000's on-chip few-shot edge-learning capability~\cite{Lutes_2024,Bratman_2023} suggests a route to user-personalised speech recognition without full retraining, which may be particularly valuable in a multi-operator industrial setting. Finally, the front-end face-landmark extractor (MediaPipe Face Mesh) currently runs on the Pi CPU and is the only non-neuromorphic stage in the pipeline. Distilling it into an Akida-compatible CNN would close the loop and make the entire perception pipeline, from raw frames to spoken-word predictions, run on the neuromorphic accelerator, removing the CPU dependency for landmark detection and consolidating the energy budget on a single device.

\section{Conclusion}
\label{sec:conclusion}
We have presented a complete neuromorphic AVSR pipeline running end-to-end on the BrainChip AKD1000 chip, with hardware-aware architectural choices, a constrained-grammar beam-search decoder and a hybrid quantization scheme tailored to the chip's deployment requirements. On the public GRID benchmark, the quantized fusion model attains 14.0\% WER under noise on the unseen-speaker split and 3.3\% WER under noise on the overlapped-speaker split. On a task-specific industrial-command corpus it reaches 98.6\% command accuracy and 1.5\% WER. On the deployment platform it sustains real-time throughput at approximately 5$\times$ less energy per inference than the strongest CPU baseline (the SNN-converted model running on the Akida software backend) and over 100$\times$ less than a laptop GPU on the video-only model. To the best of our knowledge, this is the first end-to-end multimodal AVSR system demonstrated on neuromorphic hardware of this class. The remaining gap to unconstrained SOTA accuracy is consistent with the chip's deliberate architectural simplicity, and Appendix~\ref{app:av_slower} identifies a concrete path, sparsity-aware fine-tuning, to further close the gap between the video-only and audio-video deployment points.

\appendices
\section{Why the Audio-Video Model Is Slower on the AKD1000}
\label{app:av_slower}

Table~\ref{tab:power_av} reports a substantial throughput drop for the audio-video model relative to the video-only frequency in Table~\ref{tab:power_video}. Upon inspecting the per-module hardware mapping of the audio-video checkpoint, the cause becomes clear. The image encoder, structurally identical between the video-only and audio-video pipelines, maps to 75\,NPs in 1 sequence in the video-only checkpoint but to 348\,NPs spread over 9 sequences in the audio-video checkpoint, the same footprint as the spectrogram encoder in that checkpoint (also 348\,NPs in 9 sequences). Each additional sequence corresponds to one extra hardware context switch on the AKD1000 mesh, so the remapped image encoder alone accounts for the bulk of the throughput reduction.

The root cause lies in the Akida runtime's mapping algorithm. \texttt{akida.Model.map()} is not a purely structural operation. It runs an internal binary search over \texttt{cnp\_max\_filters}, the maximum number of neurons that can share a single Convolutional Neural Processor, and at each candidate value invokes the C++ hardware-constraint solver with the model's actual weight tensors. The solver inspects each layer's non-zero weight connectivity, computed as \texttt{incoming\_conn = np.count\_nonzero(weights[\,\dots,0])}, when deciding whether the candidate split satisfies the chip's routing and bandwidth constraints. Two checkpoints with identical architecture but different trained weights therefore produce different connectivity patterns, drive the binary search to different convergence points and ultimately partition into different sequence counts. Denser or differently distributed weight matrices make the solver fall back to a smaller \texttt{cnp\_max\_filters}, which fits per-NP routing budgets only by spilling layers across additional sequences.

The training setup used in this work follows the original AkidaNet recipe, which applies $\ell_2$ weight decay only. While $\ell_2$ regularisation penalises weight magnitude, it does not drive weights to exact zero, and the mapper counts non-zeros regardless of their magnitude. As a result, the connectivity-induced asymmetry between the video-only and audio-video checkpoints persists through training. A future revision could investigate magnitude pruning of the encoder weights below a small threshold before quantization, or fine-tuning under an additional $\ell_1$ penalty. Either intervention would directly reduce \texttt{incoming\_conn} for every convolutional layer and is expected to equalise the audio-video and video-only mappings without further changes to the architecture. The mapping behaviour identified here is a generic property of the AKD1000 toolchain and therefore relevant to any work that deploys multiple checkpoints of the same architecture on this hardware.

\section{Module Layer Composition}
\label{app:layers}

This appendix details the layer composition of every module in the NAVIR pipeline. The image encoder and the spectrogram encoder reuse the AkidaNet backbone with the dense classification head removed, so we describe them by reference. The temporal-video encoder and the predictor head are NAVIR-specific and are detailed in full.

\subsection{Image Encoder and Spectrogram Encoder (AkidaNet Backbone)}
\label{app:akidanet}

Both the image and the spectrogram encoders are instances of the standard AkidaNet ImageNet model, instantiated through \texttt{akida\_models.akidanet\_imagenet} with one channel input. AkidaNet is a MobileNet-style CNN designed to be fully compatible with the AKD1000 hardware. The backbone consists of an input rescaling layer, four full convolution blocks (\texttt{conv\_0}--\texttt{conv\_3}, with strided convolutions at \texttt{conv\_0} and \texttt{conv\_2}), ten separable convolution blocks (\texttt{separable\_4}--\texttt{separable\_13}, with strided convolutions at \texttt{separable\_4}, \texttt{separable\_6} and \texttt{separable\_12}), and a global average pooling at the output of \texttt{separable\_13}. Each conv or separable block is conv $\rightarrow$ batch normalisation $\rightarrow$ ReLU. Filter counts double through the network, from 32 at \texttt{conv\_0} up to 1024 at \texttt{separable\_12} and \texttt{separable\_13}, scaled uniformly by the width multiplier $\alpha$.

In our pipeline the standard AkidaNet classification head (the dropout and 1000-way dense classifier) is removed and replaced with a single dense projection followed by a ReLU bounded at 6.0, sized to the desired embedding dimension. We use 128-dimensional per-frame embeddings for the image encoder, and 512-dimensional per-clip embeddings for the spectrogram encoder. On GRID we use $\alpha = 0.50$ for both encoders. On NAVIR we use $\alpha = 0.25$. Input shapes are 32\,$\times$\,64\,$\times$\,1 (image, GRID) and 88\,$\times$\,176\,$\times$\,1 (image, NAVIR), and 112\,$\times$\,112\,$\times$\,1 (spectrogram, both datasets). For full architectural details we refer the reader to the original AkidaNet specification distributed with the BrainChip MetaTF SDK~\cite{noauthor_overview_nodate}.

\subsection{Temporal-Video Encoder}
\label{app:videoenc}

The temporal-video encoder takes the stacked per-frame image embeddings as a 4D tensor of shape $(T, 1, F)$, where $T$ is the video window size and $F = 128$ is the per-frame embedding dimension produced by the image encoder. The width axis of size 1 lets the AKD1000's 2D convolution primitive operate as an effective 1D temporal convolution. Table~\ref{tab:videoenc} lists every layer.

\begin{table}[t]
\centering
\caption{Temporal-video encoder layers. Filter counts use the width multiplier $\alpha$ (1.0 on GRID, 0.5 on NAVIR). The output dimension $D$ is 256 on GRID and 128 on NAVIR. On NAVIR, conv blocks are conv $\rightarrow$ batch normalisation $\rightarrow$ ReLU. On GRID, batch normalisation is omitted, leaving conv $\rightarrow$ ReLU.}
\label{tab:videoenc}
\begin{tabular}{@{}llll@{}}
\toprule
Layer & Kernel & Filters & Output \\
\midrule
Input              & ---     & ---       & $(T,1,128)$ \\
Rescaling          & ---     & ---       & $(T,1,128)$ \\
\texttt{video\_conv\_0} (block) & $3\!\times\!3$ & $\lfloor 64\alpha\rfloor$  & $(T,1,\lfloor 64\alpha\rfloor)$ \\
\texttt{video\_conv\_1} (block) & $3\!\times\!3$ & $\lfloor 96\alpha\rfloor$  & $(T,1,\lfloor 96\alpha\rfloor)$ \\
\texttt{video\_conv\_2} (block) & $3\!\times\!3$ & $\lfloor 128\alpha\rfloor$ & $(T,1,\lfloor 128\alpha\rfloor)$ \\
Global avg.\ pool   & ---     & ---       & $(1,1,\lfloor 128\alpha\rfloor)$ \\
Dropout (0.03)      & ---     & ---       & $(1,1,\lfloor 128\alpha\rfloor)$ \\
\texttt{video\_dense} & ---  & $D$         & $(D)$ \\
Batch norm.\ + ReLU & ---     & ---       & $(D)$ \\
\bottomrule
\end{tabular}
\end{table}

All conv layers use \texttt{padding='same'} and stride 1, so the temporal axis is preserved through the convolution stack. Global average pooling collapses the temporal axis to produce a single video-clip embedding, which is projected to $D$ dimensions by a final dense layer with batch normalisation and bounded ReLU. The same $\ell_2$ weight decay used in AkidaNet is applied throughout.

\subsection{Predictor Head}
\label{app:head}

The predictor head is a small multilayer perceptron that consumes the concatenated $(D_v + D_s)$-dimensional fused embedding and emits per-clip token logits of dimension $V$ (the vocabulary size, including the CTC blank symbol). The depth of the head differs between the two datasets, reflecting the different complexity of their decoding tasks.

On GRID the input is a vector of dimension $D_v + D_s = 256 + 512 = 768$. The head has two hidden layers:

\begin{enumerate}
\item Dense layer with 512 units, followed by ReLU bounded at 6.0.
\item Dense layer with 256 units, followed by ReLU bounded at 6.0.
\item Dense layer with $V$ units, no activation. The output is the raw logit vector consumed by the CTC loss during training and by the constrained beam-search decoder at inference time.
\end{enumerate}

On NAVIR the input is a vector of dimension $D_v + D_s = 128 + 512 = 640$. The head uses a single hidden layer:

\begin{enumerate}
\item Dense layer with 256 units, followed by ReLU bounded at 6.0.
\item Dense layer with $V$ units, no activation.
\end{enumerate}

For modality-specific configurations the absent stream is omitted from the input. The head is quantized to 4-bit weights and activations alongside the upstream encoders, and is fine-tuned jointly with them during the QAT phase.

\printbibliography

\begin{IEEEbiographynophoto}{Leonidas Delimpasis} 
received the five-year (M.Eng. equivalent) degree in Electrical and Computer Engineering from the National Technical University of Athens, Greece, in 2025, where his diploma thesis focused on Neuro-Symbolic AI for Visual Question Answering. He has contributed to several European research projects in the areas of machine learning and computer vision. He was an ML Researcher at Tech Hive Labs. He is currently a Machine Learning Engineer at Plaixus and serves under the Hellenic National Defence General Staff, Athens, Greece. His research interests include interpretable machine learning, end-to-end development and deployment of AI systems, and AI applications in medical, industrial, and security domains.
\end{IEEEbiographynophoto}

\begin{IEEEbiography}[{\includegraphics[width=1in,height=1.25in,clip,keepaspectratio]{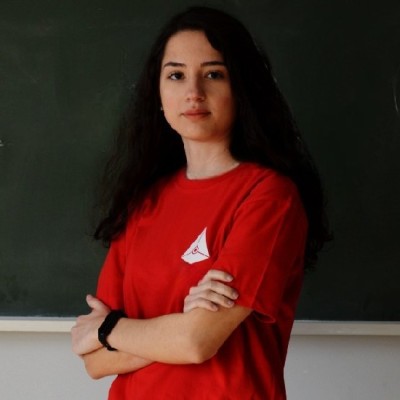}}]{Panagiota Moraiti} 
received the five-year (M.Eng. equivalent) degree in Electrical and Computer Engineering from the Democritus University of Thrace, in 2024. Her diploma thesis focused on Continual Test-Time Adaptation in the field of autonomous driving. She is currently a Computer Vision Engineer at Tech Hive Labs, where she is involved in the development of AI-driven vision systems and robotic solutions. She has contributed to both research and industrial projects. Her research interests include computer vision, deep learning and robotics. She has co-authored publications in the areas of Domain Adaptation, Continual Learning, and Automated Quality Control.
\end{IEEEbiography}

\begin{IEEEbiography}[{\includegraphics[width=1in,height=1in,clip,keepaspectratio]{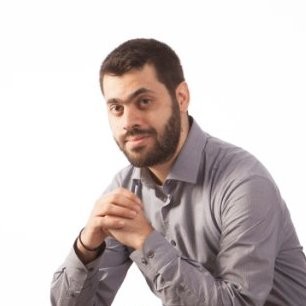}}]{Antonis Porichis} 
received the Diploma degree in Electrical and Computer Engineering from the National Technical University of Athens, Greece, in 2012, and the Ph.D. degree in Robotics and Artificial Intelligence from the University of Essex, U.K., in 2024. He is currently a Research Officer with the University of Essex. He has also held leadership roles in industry as well as research and development in data-efficient machine learning and AI-driven product innovation. Earlier in his career, he held technical leadership positions in robotics, Internet-of-Things systems, and intelligent software engineering. His research interests include machine learning, computer vision, robotic manipulation, and data-efficient learning methods, with a particular focus on imitation learning and real-world robotic applications. He has co-authored several publications in these areas, including work on robotic harvesting and learning from demonstration, published in venues such as Robotics and IEEE conferences.
\end{IEEEbiography}

\begin{IEEEbiography}[{\includegraphics[width=1in,height=1.25in,clip,keepaspectratio]{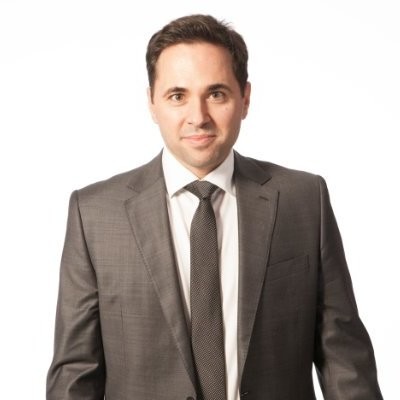}}]{Panos Chatzakos} 
received the Diploma, M.Sc., and Ph.D. degrees in Mechanical Engineering from the National Technical University of Athens, Greece. He is currently the Director of the Essex Artificial Intelligence Innovation Centre at the University of Essex, U.K., a joint initiative with TWI Ltd., where he leads research and technology transfer activities in artificial intelligence and robotics. He has extensive experience in both academia and industry, having established and led technology organizations and innovation initiatives across Europe, including senior leadership roles at TWI and as Executive Chairman of Tech Hive Labs. He has successfully managed and delivered numerous large-scale research and innovation projects, collaborating with industrial partners, SMEs, and academic institutions, and has led multidisciplinary teams developing AI-driven and robotic systems for real-world applications. His work spans data science, robotics, and intelligent systems integration, with a strong focus on translating research into commercially viable technologies. Dr. Chatzakos has authored and co-authored several scientific publications and has received multiple international awards for his contributions to robotics and applied research, including distinctions in industrial robotics and advanced manufacturing.
\end{IEEEbiography}

\begin{IEEEbiography}[{\includegraphics[width=1in,height=1.25in,clip,keepaspectratio]{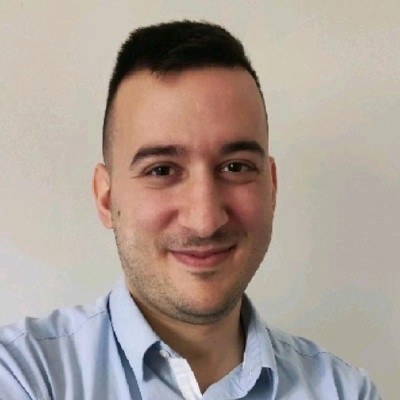}}]{Michail Karamousadakis} 
received the five-year (M.Sc. equivalent) degree in Electrical and Computer Engineering from the National Technical University of Athens, Greece, in 2019, and the M.Sc. degree in Rehabilitation Engineering from the National and Kapodistrian University of Athens, Greece, in 2023. He has been involved in the development of AI-driven systems and edge computing solutions. He has also held software engineering and research roles, contributing to the design and implementation of intelligent systems and cyber-physical applications. His research interests include artificial intelligence, machine learning, edge computing, DevOps/MLOps, and robotics, with a focus on deploying scalable and efficient AI systems in real-world environments. He has co-authored several publications in these areas, including works on edge AI systems, cyber-physical security, and data-driven intelligent platforms.
\end{IEEEbiography}

\EOD

\end{document}